\documentclass[letterpaper]{article} 

\usepackage{aaai2026}  
\usepackage{times}  
\usepackage{helvet}  
\usepackage{courier}  
\usepackage[hyphens]{url}  
\usepackage{graphicx} 
\usepackage{booktabs}
\usepackage{multirow}
\usepackage{makecell}
\usepackage{amsfonts}
\usepackage{natbib}  
\usepackage{caption} 
\usepackage{algorithm}
\usepackage{algorithmic}

\usepackage{newfloat}
\usepackage{listings}
\DeclareCaptionStyle{ruled}{labelfont=normalfont,labelsep=colon,strut=off} 
\floatstyle{ruled}
\newfloat{listing}{tb}{lst}{}
\floatname{listing}{Listing}
\title{Motion is the Choreographer: Learning Latent Pose Dynamics \\for Seamless Sign Language Generation}
\author{
    Jiayi He, Xu Wang, Shengeng Tang, Yaxiong Wang, Lechao Cheng, Dan Guo
}
\affiliations{
    School of Computer Science and Information Engineering, Hefei University of Technology\\

    \{hejy4396, wangxu2002\}@mail.hfut.edu.cn, \{tangsg, wangyx, chenglc, guodan\}@hfut.edu.cn
}

\usepackage{bibentry}

\begin{document}
\nocopyright
\maketitle
\nocopyright
\begin{abstract}

Sign language video generation requires producing natural signing motions with realistic appearances under precise semantic control, yet faces two critical challenges: excessive signer-specific data requirements and poor generalization. We propose a new paradigm for sign language video generation that decouples motion semantics from signer identity through a two-phase synthesis framework. First, we construct a signer-independent multimodal motion lexicon, where each gloss is stored as identity-agnostic pose, gesture, and 3D mesh sequences, requiring only one recording per sign. This compact representation enables our second key innovation: a discrete-to-continuous motion synthesis stage that transforms retrieved gloss sequences into temporally coherent motion trajectories, followed by identity-aware neural rendering to produce photorealistic videos of arbitrary signers. Unlike prior work constrained by signer-specific datasets, our method treats motion as a first-class citizen: the learned latent pose dynamics serve as a portable "choreography layer" that can be visually realized through different human appearances. Extensive experiments demonstrate that disentangling motion from identity is not just viable but advantageous - enabling both high-quality synthesis and unprecedented flexibility in signer personalization. 
\end{abstract}

\section{Introduction}
According to the World Federation of the Deaf, there are more than 70 million deaf people worldwide. This large group relies mainly on sign language as the main means of communication, with more than 300 distinct sign languages in global use. These sign languages are natural languages, structurally independent of spoken languages, and embody unique linguistic identities and cultural diversity. Recently, with the rapid development of deep learning, Sign Language Generation \& Synthesis (SLG) has gradually become an extremely challenging emerging research field.

\begin{figure}[t]
  \centering
  \includegraphics[width=1.0\columnwidth]{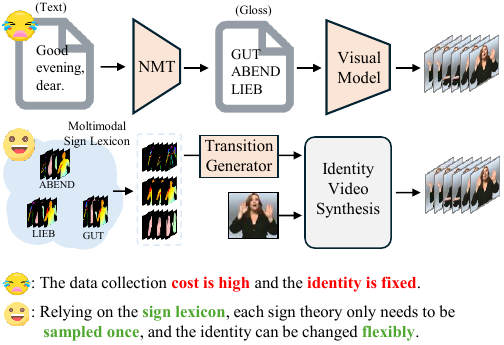}
  \caption{\textbf{Top:} Traditional SLG methods employ a two-stage process: first utilizing NMT to convert text into gloss, followed by visual models to synthesize sign videos. \textbf{Bottom:} Our method employs three key components: (1) a multimodal sign lexicon for motion representation, (2) a transition generator to decouple motion semantics from signer identity, and (3) an identity video synthesis module. We enable flexible identity transformation through transition generation while significantly reducing data requirements.}
  \label{Motivation}
\end{figure}

SLG aims to establish an accurate mapping between spoken and sign languages. As shown at the top of Figure~\ref{Motivation}, conventional SLG methods~\cite{stoll2020text2sign, stoll2020signsynth} typically follow this pipeline: text-to-gloss (T2G) translation and gloss-to-video generation. While Neural Machine Translation (NMT)~\cite{othman2011statistical} has proven effective for T2G conversion, current research primarily focuses on the generation stage. Existing methods generally fall into two categories: autoregressive~\cite{saunders2020progressive, wangxu2025, tang2022gloss, yin2024t2s} and non-autoregressive~\cite{xie2024g2p, tang2025sign, tang2025gloss} approaches. A key limitation of these methods is their dependence on large-scale annotated sign language sentence datasets.

To this end, a scheme was proposed to first collect discrete sign language units and then combine them into sentences. Early work~\cite{wang2002method, sagawa2002teaching} simply spliced discrete sign language units together, resulting in impaired visual fluency. Sign-D2C~\cite{tang2025discrete} attempts to synthesize discrete sign poses into continuous sentences, but due to the lack of a dataset specifically for synthesizing smooth transitions, it only verifies the effectiveness of the method through a fixed masking method, and the reliability of the conclusions is limited. In contrast, our method leverages continuous sign language recognition (CSLR)~\cite{chen2022two}  to split discrete sign language units from the complete sentence dataset, which enables effective evaluation of the proposed method and thus more comprehensively verifies its performance.

In this work, we propose MicT, a novel pose-centric framework that decomposes sign language video synthesis into three interconnected stages. First, we construct a multimodal sign lexicon by segmenting continuous sign sequences into discrete units and extracting fine-grained 3D pose representations using state-of-the-art tools (DWPose~\cite{yang2023effective}, HaMeR~\cite{pavlakos2024reconstructing}, and SMPLer-X~\cite{cai2023smpler}). Second, our transformation generator module learns to model 3D motion trajectories, facilitating smooth spatiotemporal transitions between isolated signs. By capturing the intrinsic dynamics of sign language, the module addresses motion discontinuity issues while preserving natural motions. Finally, the video synthesis module renders realistic videos from coherent motions while maintaining target signer identity, effectively bridging abstract motion trajectories with concrete visual manifestations. Our framework explicitly models gesture dynamics as a continuous flow, inherently supporting interpolation of novel vocabulary combinations. This pose-centric design establishes motion as the fundamental orchestrator of synthesis while maintaining adaptability to diverse signers and semantic contexts.


In summary, our main contributions are as follows:
\begin{itemize}
\item We propose a paradigm shift from end-to-end generation to a two-phase motion-first pipeline (discrete-to-continuous motion synthesis + identity-aware rendering), eliminating the need for full-sentence videos or signer-specific datasets. Theoretical analysis confirms our approach's generalizability to novel sign combinations through vocabulary recombination.
 
\item We introduce an identity-decoupled representation scheme that stores each sign as portable motion primitives (pose/gesture/mesh sequences), requiring only single recordings per gloss. This achieves more greater data efficiency than conventional approaches while enabling arbitrary signer personalization.

\item The proposed method establishes new benchmarks on PHOENIX14T, outperforming previous approaches in both motion quality and visual realism. The results validate that motion-centric representation enables superior semantic accuracy while maintaining appearance fidelity.
\end{itemize}

\begin{figure*}[tbh]
  \centering
  \includegraphics[width=1.0\textwidth]{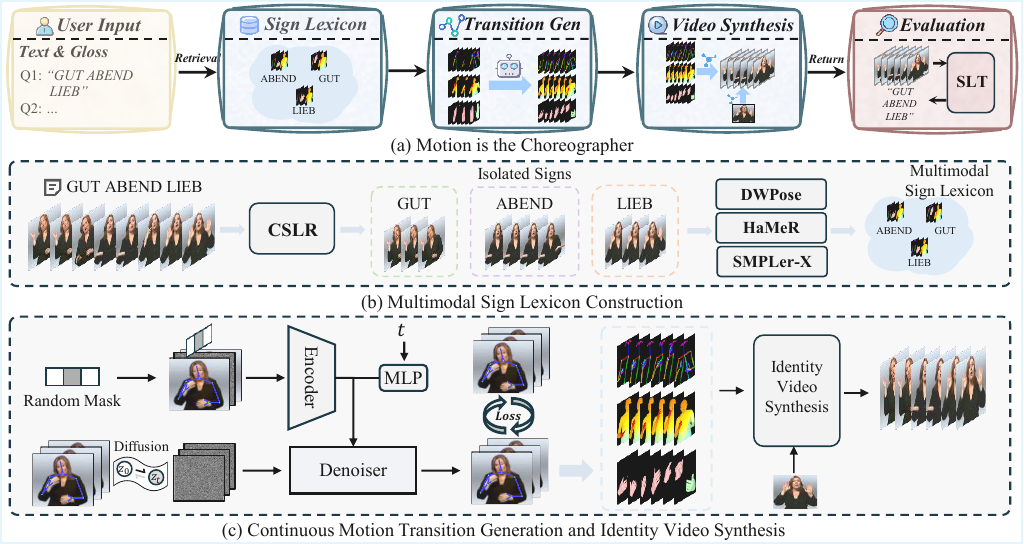}
  \caption{Overview of the sign language video generation framework. (a)MicT's pipeline. (b) Multimodal sign lexicon construction:
  We use the TwoStream-SLR model~\cite{chen2022two} to segment the continuous sign language into a set of isolated signs, and then use DWPose~\cite{yang2023effective}, Hamer~\cite{pavlakos2024reconstructing} and Smpler-x~\cite{cai2023smpler} to extract the pose, hand details and 3D full-body and store them in the multimodal sign language lexicon. (c) Continuous Motion Transition Generation and Identity Video Synthesis: We achieve continuous sign language synthesis based on multimodal sign language motions by generating smooth transitions between motions, and use identity-aware video synthesis to generate sign language videos that match the characteristics of a specified identity.}
  \label{fig: Stage1 and Stage2}
\end{figure*}

\section{Related Work}
\subsection{Sign Language Generation \& Synthesis}
Sign language is a primary mode of communication for the deaf and hard-of-hearing community. Advances in artificial intelligence have driven rapid developments in sign language research. The current research primarily aims to construct a bidirectional system through sign language translation (SLT)~\cite{tangslt2022, Guo_Tang_Wang, gong2024llms, Chen_2022_CVPR}, sign language recognition (SLR)~\cite{xue2023alleviating, zhao2023best, ahn2024slowfast}, and sign language generation (SLG). However, the level of development of SLG is not yet sufficient to support the achievement of this goal.

Early SLG efforts involved building sign language databases and then synthesizing sign language sentences using animation technology~\cite{Mazumder_Mukhopadhyay_Namboodiri_Jawahar_2021, McDonald_Wolfe_Schnepp_Hochgesang_Jamrozik_Stumbo_Berke_Bialek_Thomas_2016, Segouat_2009}. However, these technologies were limited by high collection costs and were restricted to specific terminology. With the advent of Transformer~\cite{vaswani2017attention}, SLG has achieved a significant breakthrough. Saunders \emph{et al.} proposed the first transformer model for generating sign language pose~\cite{saunders2020progressive} and conducted a series of extensions based on this model~\cite{saunders2021continuous}. In response to the inherent characteristics of sign language, Tang, Wang, and others proposed a series of diffusion model-based solutions~\cite{tang2025sign, tang2025gloss, xie2024g2p}, which were guided by semantics.

Although the above methods have made significant progress, the scarcity of sign language data remains a significant limitation. Zuo \emph{et al.} proposed a baseline method for generating sign language 3D avatars: using an existing sign language recognition method~\cite{chen2022two} to segment sign language movements in sentence datasets and store them in the dataset, retrieving corresponding sign language movements under given text conditions, and then synthesizing sign language 3D avatars. However, their method is still limited by text and characters. In contrast, our method can generate sign language videos without the constraints of text and characters.

\subsection{Diffusion-based Generative Models}
Diffusion models were first introduced by Sohl-Dickstein \emph{et al.}~\cite{Sohl-Dickstein_Weiss_Maheswaranathan_Ganguli_2015}, comprising two fundamental stages: a forward process and a reverse process. The forward process gradually corrupts the original data through iterative Gaussian noise addition until it converges to pure noise. The reverse process employs a neural network to learn the data distribution and progressively denoise random samples, ultimately generating novel samples that match the training data distribution. While early diffusion models suffered from computational inefficiency and suboptimal generation quality, Ho \emph{et al.}'s DDPM~\cite{ho2020denoising} made significant advances by streamlining the sampling process and optimizing the network architecture. This breakthrough marked the beginning of rapid development in diffusion models, leading to numerous influential variants~\cite{rombach2022high, gu2022vector, jia2025d, peebles2023scalable, Song_Meng_Ermon_2020}.

Diffusion models have demonstrated exceptional generative capabilities, producing both diverse and high-quality outputs across multiple domains due to their inherent flexibility. Their impact is particularly notable in: (1) image generation~\cite{jia2025d, peebles2023scalable}, where they achieve state-of-the-art fidelity; (2) super-resolution~\cite{zhang2025uncertainty, yue2025arbitrary}, enabling high-quality resolution enhancement; and (3) video generation~\cite{wang2025lavie, zhou2024realisdanceequipcontrollablecharacter}, facilitating realistic motion synthesis. Within Sign Language Generation (SLG), diffusion models have inspired several key approaches: GCDM~\cite{tang2025gloss} employs multi-hypothesis generation for gesture synthesis, G2P-DDM~\cite{xie2024g2p} formulates the task as discrete prediction, and Sign-IDD~\cite{tang2025sign} introduces 4D action representations for improved accuracy. However, these methods all require complete sentence-level datasets. While Sign-D2C~\cite{tang2025discrete} attempts continuous synthesis from discrete poses, its evaluation remains limited to masked scenarios due to dataset constraints, leaving its generalizability uncertain. In our work, we go beyond simple pose transition generation. We use an advanced CSLR model to divide continuous sign language sentences into discrete sign language words, build a multimodal sign language database based on this, and then perform transition generation. This allows us to truly evaluate the generation effect of sign language transition.

\section{Methodology}
\subsection{Preliminary}
We propose an innovative sign language video generation scheme that decouples the traditional end-to-end sentence-to-video generation process into three stages: vocabulary matching, multimodal transition generation, and identity decoupling synthesis. This successfully breaks through the dependence of existing methods on fully annotated video data. Our method leverages a multimodal sign lexicon to generate multi-identity sign language videos for arbitrary novel sentence combinations, requiring only a single recording per vocabulary item in theory. In contrast to existing methods~\cite{zuo2024simple, tang2025sign}, our method is more flexible and efficient.

\subsection{Multimodal Sign Lexicon Construction}
As shown in Figure~\ref{fig: Stage1 and Stage2} (a), our method is built upon a multimodal sign lexicon comprising paired data consisting of text, video, poses (\textbf{DWPose}), hand details (\textbf{HaMeR}) and 3D full-body (\textbf{SMPLer-X}). However, the existing PHOENIX14T~\cite{Camgoz_Hadfield_Koller_Ney_Bowden_2018} dataset is a sentence dataset and does not provide the other three modalities besides video. To build a multimodal sign lexicon, we were inspired by Spoken2Sign~\cite{zuo2024simple} and used the TwoStream-SLR~\cite{chen2022two} to divide continuous sign sentences into discrete signs. On this basis, we use DWPose~\cite{yang2023effective}, HaMeR~\cite{pavlakos2024reconstructing}, and SMPLer-X~\cite{cai2023smpler} to extract poses, hand details, and 3D full-body to form a multimodal sign lexicon.

\subsection{Continuous Motion Transition Generation}
Given an input sentence, we first segment it into glosses and retrieve their corresponding multimodal sequences from the lexicon (Sec. \textbf{3.2 Multimodal Sign Lexicon Construction}). Due to the differences between pose and hand details (\textbf{HaMeR}), 3D full-body (\textbf{SMPLer-X}), we uniformly process them into a $1D$ data format and refer to them collectively as motion in the following.
There are $N$ glosses corresponding to sign \textsl{Y} = \{$y_1$, $y_1$, \dots, $y_N$\}, which together form the sign sequence $n$ of $t$ frames. Our goal is to generate transition motions between adjacent $N$ signs. We add $x$ frames between $N$ signs as transition frames to be generated to form the observation conditions $m$ of sign motions. We use an embedding layer to map the observed motion into a high-dimensional space. The embedding layer consists of a linear layer and a position embedding layer. The formula is as follows:
\begin{eqnarray}
    m' = W^{m} \cdot PE(m) + b^{m},
\end{eqnarray}
where $PE$ is implemented by a predefined sine function, and $W^{m}$ and $b^{m}$ are both learnable parameters.

Then, $m'$ passes through the Multi-Head Attention layer and the Feed-Forward layer to obtain the observed feature $m_{c}'$. The feature extraction process can be expressed as:
\begin{eqnarray}
    m_{c}' = FFL(MHA(NL(m')) + m'),
\end{eqnarray}
where, $FFL$ is the Feed Forward layer, $MHA$ is the Multi-Head Attention layer, and $NL$ is the normalization layer.

Specifically, $MHA$ plays an important role in sequentially processing the contextual dependencies of $m'$. As well as known, $MHA$ projects the input into \textit{Query-Q}, \textit{Key-K}, and \textit{Value-V} through linear matrices as follows:
\begin{eqnarray}
    Q = m' \cdot W^{m}_{Q} \qquad   K = m' \cdot W^{m}_{K} \qquad  V = m' \cdot W^{m}_{V}.
\end{eqnarray}
The $MHA$ formula can be expressed as:
\begin{eqnarray}
    Attention(\textit{Q}, \textit{K}, \textit{V}) = Softmax(\frac{QK^{T}}{\sqrt{d}})\textit{V},
\end{eqnarray}
where $d$ is the scaling factor.

During the training phase, after obtaining the encoded observation condition $m_{c}'$, we add noise to the complete sign sequence $n$. The formula is as follows:
\begin{eqnarray}
    n_t = \alpha_t \cdot n_{t-1} + \sqrt{1 - \alpha_t^2} \cdot \epsilon_t,
\end{eqnarray}
where $\alpha_t$ is generated by the cosine scheduler of the diffusion model and $\epsilon_t$ is sampled from Gaussian noise.

Our $Denoiser$ is also built using $MHA$ and conditionally embedded via $Cross-Attention$, where $m_{c}'$ generates, \textit{Key-K}, \textit{Value-V}. Finally, we use the denoiser to recover the uncontaminated sign sequence from $n_t$ given the observation $m_{c}'$. The formula is as follows: 
\begin{eqnarray}
     \widetilde{n} = Denoiser(n_t, m_{c}', t),
\end{eqnarray}
where $t$ represents the sampling step of diffusion.

During inference, we generate complete sign sequences $\widetilde{n}$ by sampling from Gaussian noise $n_t$ given the available observations $m_{c}'$. The formula is as follows: 
\begin{eqnarray}
\left\{\begin{array}{l}
    n_{t-1} = \sqrt{\bar{a}_{t-1}} \cdot \tilde{n_0} + \sqrt{1-\bar{a}_{t-1}-\sigma_t^2} \cdot \epsilon_t + \sigma_t\epsilon,\\
    \epsilon_t = ({{n_{t}}}-\sqrt{\bar{a}_{t}} \cdot {{\tilde{n_0}}}) / \sqrt{1-\bar{a}_{t}},\\
     \sigma_t = \sqrt{(1-\bar{a}_{t-1})/(1-\bar{a}_{t})} \cdot \sqrt{1-\bar{a}_{t}/\bar{a}_{t-1}},
\end{array}\right.
\end{eqnarray}

In this stage, $n_{t-1}$ is utilized as input of $\mathcal{D}$ to regenerate and update ${p_0}$, which is repeated $I$ times. Initiated at $T$, the timestep for each iteration is computed as $T = T - (1 - i / I)$, where $i \in [0, I]$. The adjustable parameter $i$ controls the diversity and quality of the generated results. 

\begin{figure}[t]
  \centering
  \includegraphics[width=1.0\columnwidth]{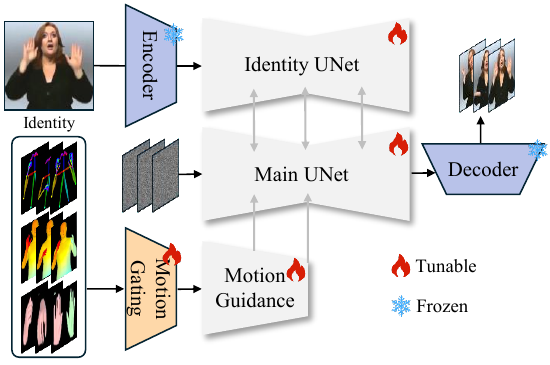}
  \caption{Architecture of identity-aware video synthesis. Given a complete sign language motion $\widetilde{m}$ and an identity, synthesize identity-aware videos.}
  \label{fig: Stage3}
\end{figure}

\subsection{Identity-Aware Video Synthesis}

Our identity-aware video synthesis stage is designed as a modular framework that can flexibly incorporate different rendering backends. The architecture, illustrated in Figure 3, establishes three core requirements for any implementation: (1) motion-conditioned generation capability to maintain signing accuracy, (2) robust identity preservation mechanisms, and (3) multimodal error correction to handle potential discrepancies between pose, gesture, and mesh inputs. In this work, we adopt RealisDance~\cite{zhou2024realisdanceequipcontrollablecharacter} as it satisfies these requirements while demonstrating state-of-the-art performance in pose-to-video generation tasks. However, the framework is intentionally designed to be model-agnostic, allowing for future integration of more advanced rendering techniques as the field progresses.

\noindent\textbf{Multimodality.} 
RealisDance supports pose, hand details and 3D full body in video format as input, which greatly reduces the threshold for use. The multimodal motion rendering process employs two key components that are fundamental: a posture gating module that dynamically weights and balances the different motion modality inputs (pose, hands, and 3D mesh), and cross-modal alignment mechanisms that resolve potential discrepancies between these modalities. This design ensures robust video generation regardless of the specific backend implementation, addressing a critical challenge in sign language synthesis where different motion representations may contain slight variations.

\begin{table*}[!htbp]
\renewcommand\arraystretch{1.1}   
\centering
\resizebox{\textwidth}{!}{
\begin{tabular}{cccccccccccc}
\toprule[1pt]
\multirow{2}{*}{Methods}&\multicolumn{5}{c}{DEV} & ~ &\multicolumn{5}{c}{TEST} \\
\cline{2-6} \cline{8-12}
~ &BLEU-1$\uparrow$ &BLEU-4$\uparrow$ &ROUGE$\uparrow$ &SSIM$\uparrow$ &PSNR$\uparrow$  & ~ &BLEU-1$\uparrow$ &BLEU-4$\uparrow$ &ROUGE$\uparrow$ &SSIM$\uparrow$ &PSNR$\uparrow$\\ 
\midrule[0.5pt]
\multicolumn{1}{c}{Ground Truth} & 34.42 & 15.57 & 35.07 & - & - & ~ & 34.82 & 15.68 & 34.77 & - & - \\ 
\midrule[0.5pt]
\multicolumn{1}{c}{PTSLP~\cite{saunders2020progressive}}   & 13.75 & 4.42 & 13.67 & \underline{0.95} & 19.04 &~ & 13.26 & 4.32 & 13.03 & \underline{0.95} & 19.06 \\ 
\multicolumn{1}{c}{CogvideoX~\cite{yang2024cogvideox}}  & 15.42 & 4.89 & 14.67 & 0.92 & 14.92 &~  & 15.16 & 5.62 & 14.17 & 0.92 & 14.93 \\ 
\multicolumn{1}{c}{SignAligner~\cite{wang2025signaligner}}  & \underline{30.09} & \underline{12.19} & \underline{29.89} & \bf{0.96} & \underline{21.32}  &~ & \underline{29.94} & \underline{12.27} & \underline{29.12} & \bf{0.96} & \underline{21.31} \\ 
\midrule[0.5pt]
\multicolumn{1}{c}{Ours} & \bf{31.57} & \bf{13.35} & \bf{31.62} & \bf{0.96} & \bf{22.64} &~  & \bf{30.53} & \bf{14.29} & \bf{30.49} & \bf{0.96} & \bf{22.55} \\
\bottomrule[1pt]
\end{tabular}}
\caption{Results on the PHOENIX14T dataset for Text to hand details tasks.}
\label{ham}
\end{table*}

\begin{table*}[!htbp]
\renewcommand\arraystretch{1.1}   
\centering
\label{tab: ham and smp}
\resizebox{\textwidth}{!}{
\begin{tabular}{cccccccccccc}
\toprule[1pt]
\multirow{2}{*}{Methods}&\multicolumn{5}{c}{DEV} & ~ &\multicolumn{5}{c}{TEST} \\
\cline{2-6} \cline{8-12}
~ &BLEU-1$\uparrow$ &BLEU-4$\uparrow$ &ROUGE$\uparrow$ &SSIM$\uparrow$ &PSNR$\uparrow$  & ~ &BLEU-1$\uparrow$ &BLEU-4$\uparrow$ &ROUGE$\uparrow$ &SSIM$\uparrow$ &PSNR$\uparrow$\\ 
\midrule[0.5pt]
\multicolumn{1}{c}{Ground Truth} & 31.95 & 13.89 & 31.54 & - & - & ~ & 32.78 & 14.17 & 32.20 & - & - \\ 
\midrule[0.5pt]
\multicolumn{1}{c}{PTSLP~\cite{saunders2020progressive}}  & 10.35 & 2.19 & 10.25 & \underline{0.79} & 16.10 &~ & 9.89 & 2.62 & 9.65 & 0.79 & 16.23 \\ 
\multicolumn{1}{c}{CogvideoX~\cite{yang2024cogvideox}}  & 13.31 & 1.50 & 11.05 & 0.71 & 12.05 &~ & 9.85 & 1.56 & 7.67 & 0.72 & 12.03 \\ 
\multicolumn{1}{c}{SignAligner~\cite{wang2025signaligner}}  & \underline{25.42} & \underline{10.04} & \underline{26.00} & \underline{0.83} & \underline{18.62}  &~ & \underline{27.48} & \underline{11.20} & \underline{27.00} & \underline{0.83} & \underline{18.65} \\ 
\midrule[0.5pt]
\multicolumn{1}{c}{Ours} & \bf{29.83} & \bf{12.42} & \bf{30.18} & \bf{0.86} & \bf{21.61} &~ & \bf{30.72} & \bf{13.25} & \bf{30.26} & \bf{0.86} & \bf{21.52} \\
\bottomrule[1pt]
\end{tabular}}
\caption{Results on the PHOENIX14T dataset for Text to 3D full body tasks.}
\label{smp}
\end{table*}

\begin{table*}[!htbp]
\renewcommand\arraystretch{1.1}   
\centering
\resizebox{\textwidth}{!}{
\begin{tabular}{ccccccccccccc}
\toprule[1pt]
\multirow{2}{*}{Methods}&\multicolumn{5}{c}{DEV} & ~ &\multicolumn{5}{c}{TEST} \\
\cline{2-6} \cline{8-12}
~ &BLEU-1$\uparrow$ &BLEU-4$\uparrow$ &ROUGE$\uparrow$ &SSIM$\uparrow$ &PSNR$\uparrow$ & ~ &BLEU-1$\uparrow$ &BLEU-4$\uparrow$ &ROUGE$\uparrow$ &SSIM$\uparrow$ &PSNR$\uparrow$ \\ 
\midrule[0.5pt]
\multicolumn{1}{c}{Ground Truth}  & 38.38 & 16.95 & 37.56 & - & - &~ & 38.45 & 17.42 & 37.65 & - & - \\ 
\midrule[0.5pt]
\multicolumn{1}{c}{PTSLP + RealisDance~\cite{zhou2024realisdanceequipcontrollablecharacter}}  & 8.55 & 1.68 & 9.15 & \underline{0.58} & 11.28 &~ & 8.86 & 1.52 & 8.83 & \underline{0.58} & 11.45 \\ 
\multicolumn{1}{c}{CogvideoX~\cite{yang2024cogvideox}}  & 8.14 & 0.46 & 7.21 &  0.29 & 3.82 &~ & 8.40 & 0.51 & 7.33 & 0.29 & 3.82 \\ 
\multicolumn{1}{c}{SignAligner~\cite{wang2025signaligner}}  & \underline{19.33} & \underline{7.36} & \underline{21.08} & \bf{0.73} & \underline{15.29}  &~ & \underline{20.56} &  \underline{8.17} & \underline{20.88} & \bf{0.73} & \underline{15.32} \\ 
\midrule[0.5pt]
\multicolumn{1}{c}{Ours} & \bf{20.70} & \bf{8.45} & \bf{22.58} & \bf{0.73} & \bf{15.81} &~ & \bf{21.17} & \bf{8.55} & \bf{22.45} & \bf{0.73}  & \bf{15.93} \\
\bottomrule[1pt]
\end{tabular}}
\label{tab: video}
\caption{Results on the PHOENIX14T dataset for sign language video synthesis tasks.}
\end{table*}

\begin{table}[tbp]
\renewcommand\arraystretch{1.1}   
\centering
\resizebox{\columnwidth}{!}{
\begin{tabular}{ccccc}
\Xhline{1pt}
Methods & BLEU-1$\uparrow$ & BLEU-4 $\uparrow$& ROUGE $\uparrow$ & WER $\downarrow$\\
\Xhline{0.5pt}
PTSLP & 13.35 & 4.31 & 13.17 & 96.50\\
NAT-AT & 14.26 & 5.53 & 18.72 & 88.15\\
NAT-ET & 15.12 & 6.66 & 19.43 & 82.01\\
DET & 17.18 & 5.76 & 17.64 & -\\
G2P-DDM & 16.11 & 7.50 & - & 77.26\\
GCDM & 22.03 & 7.91 & 23.20 & 81.94\\
GEN-OBT & 23.08 & 8.01 & 23.49 & 81.78\\
SignAligner & \underline{24.39} & \underline{8.47} & \bf{25.21} & \underline{73.89}\\
\Xhline{0.5pt}
Ours & \bf{24.62} & \bf{9.11} & \underline{24.68} & \bf{73.00} \\
\Xhline{1pt}
\end{tabular}}
\caption{Results on PHOENIX14T for Text to Pose task.}
\label{t2p}
\end{table}

\noindent\textbf{Identity awareness.} 
For identity-conditioned generation, our method implements hierarchical identity injection at both global and local levels. The system maintains a strict motion-identity disentanglement that prevents attribute leakage during generation. Here, identity awareness achieves high consistency of different reference character identities through the Identity UNet module. When generating videos, it can accurately capture the character features (such as clothing, appearance, etc.) in the identity image and ensure that each frame of the generated video strictly follows these features, thereby avoiding identity drift or distortion. Our motion-first pipeline can produce high-quality, identity-preserving sign language videos when paired with an appropriate rendering system. 

\subsection{Training}
\noindent\textbf{Continuous motion transition generation.} 
In the Continuous Motion Transition Generation stage, we use the following method to generate continuous motion sequences: First, we simulate incomplete motion input by applying a random mask mechanism (with a probability of 0.3) on the complete sentence dataset to construct observation conditions. Based on these conditions, we use a diffusion model for motion completion: 1) add progressive noise to the complete motion sentence to construct a diffusion process; 2) under the constraint of known partial motion (\emph{i.e.}, observation conditions), gradually restore the uncontaminated complete motion sequence through an inverse denoising process. This method effectively achieves natural motion transition generation under partial observation conditions. Following~\cite{huang2021towards, saunders2020progressive}, we adopt a joint loss to constrain the accuracy of the joint positions in poses, ensuring precise matching with the ground truth. The joint constraint $\mathcal{L}_{joint}$ is defined as follows: 
\begin{eqnarray}
    \mathcal L_{joint} = \frac{1}{S}\sum_{s=1}^{S}|p_s-\hat{p_s}|,
\end{eqnarray}
where $p^s_0$ and $\hat{p^s_0}$ represent the ground truth and generated 3D pose at the $s$-th frame, respectively.

\begin{figure*}[tbh]
  \centering
  \includegraphics[width=1.0\textwidth]{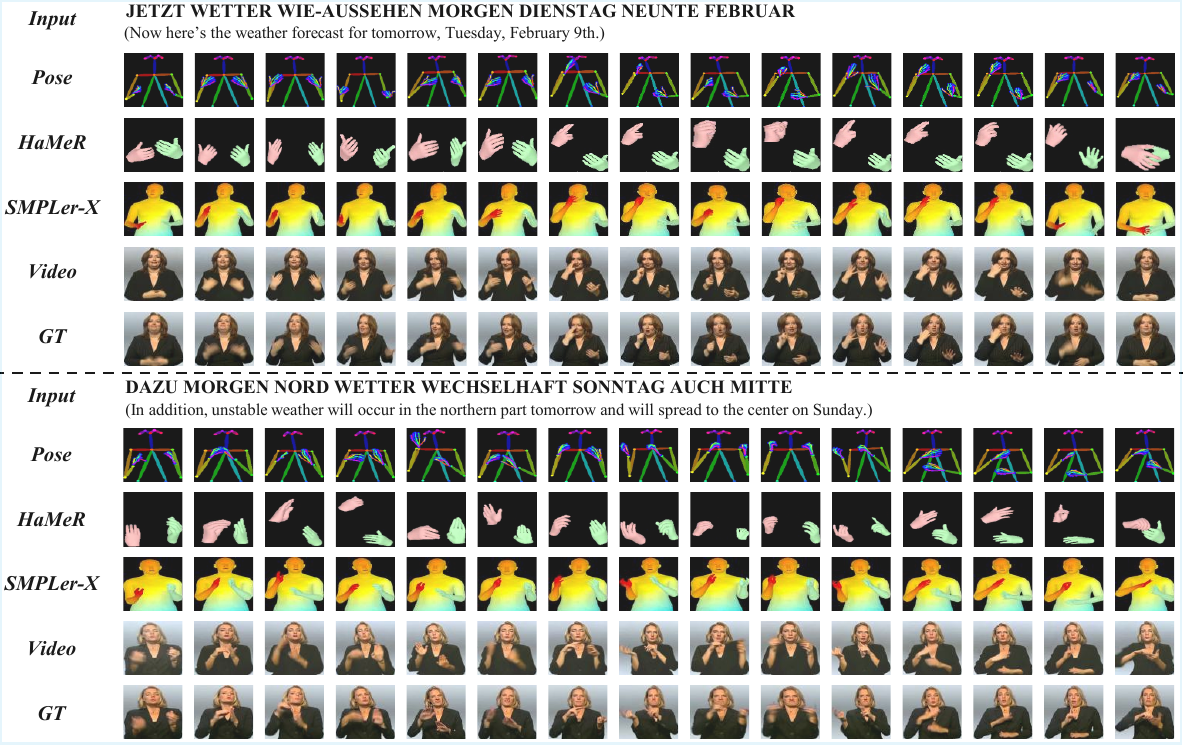}
  \caption{Visualization examples of generated videos, poses, hand details, and 3D full-body on PHOENIX14T.}
  \label{video}
\end{figure*}

\noindent\textbf{Photorealistic sign video rendering.}
RealisDance adopts a two-stage training strategy to efficiently fine-tune specific components while freezing key modules. The first stage fine-tunes images based on the Real Vision v5.1 pre-trained model, freezes DINOv2 and motion modules, and focuses on optimizing the main UNet and reference UNet to maintain character consistency; the second stage initializes the motion module from AnimateDiff, freezes the image generation part, specifically trains timing-related components, and improves robustness through random posture enhancement.

\section{Experiments}
\subsection{Experimental Setup}
\noindent\textbf{Dataset.}
We experimentally evaluate our approach on the publicly available German Sign Language corpus, the RWTH-PHOENIX-Weather2014T (PHOENIX14T) dataset~\cite{Camgoz_Hadfield_Koller_Ney_Bowden_2018}. This dataset contains 1085 glosses and 8257 sign language samples.

\noindent\textbf{Comparison Methods}
We compare the pose sequences generated by our method through continuous motion transition generation with the mainstream G2P methods in the SLP. For hand detail modeling and 3D full body pose generation, due to the lack of existing benchmarks, we build a comparative baseline by fully tuning the PTSLP~\cite{saunders2020progressive} and CogvideoX~\cite{yang2024cogvideox}.

\noindent\textbf{Evaluation metrics.} Based on previous work, we use NSLT~\cite{camgoz2018neural} and GFSLT~\cite{zhou2023gloss} as back-translation models for continuous motion transition generation and video synthesis, respectively. We mainly use \textit{BLEU}, \textit{ROUGE}, and \textit{WER} to evaluate the semantic accuracy of our method. In addition, we report the final video quality through \textit{SSIM}, \textit{PSNR}, and \textit{FID}. 

\noindent\textbf{Implementation details.} Our multimodal sign lexicon is constructed mainly using DWPose~\cite{yang2023effective}, HaMeR~\cite{pavlakos2024reconstructing} and SMPLer-X~\cite{cai2023smpler}.

DWPose: Responsible for capturing basic whole-body movements, providing 2D key point sequences of the human body, hands, and face as the backbone control signal for overall movement.

HaMeR: Focuses on fine hand gestures, accurately models finger joint angles and spatial positions through 3D mesh reconstruction, and ensures the authenticity of complex gestures.

SMPLer-X: Provides 3D human structure and semantic information, encodes body shape parameters, depth, and continuous semantics into a unified color surface representation, and enhances the model's understanding of limb spatial relationships.

\noindent\textbf{Model Parameters} The DDIM~\cite{Song_Meng_Ermon_2020} used in our Continuous Motion Transition Generation stage uses a $Cosine$ noise scheduler, with a noise addition time $T$ of $1000$ and a ddim sampling step $i$ of $5$. Our Encoder and Denoiser both use Transformer with an embedding dimension of $512$, 2 attention layers and 4 heads. We use PyTorch with the Adam optimizer~\cite{Kingma_Ba_2014} on an NVIDIA GeForce RTX A40 GPU with a learning rate of $1\times10^{-3}$.

\subsection{Comparison}
\noindent\textbf{Smooth motion transitions.} As shown in Table~\ref{t2p}, our method surpasses the existing methods in some key indicators: BLEU-1 reaches 24.62 and BLEU-4 is 9.11. Compared with previous methods (such as SignAligner~\cite{wang2025signaligner} and GEN-OBT~\cite{tang2022gloss}), our model performs better, which shows that it has a significant advantage in generating accurate and coherent pose sequences from text descriptions.

\begin{figure}[t]
  \centering
  \includegraphics[width=1.0\columnwidth]{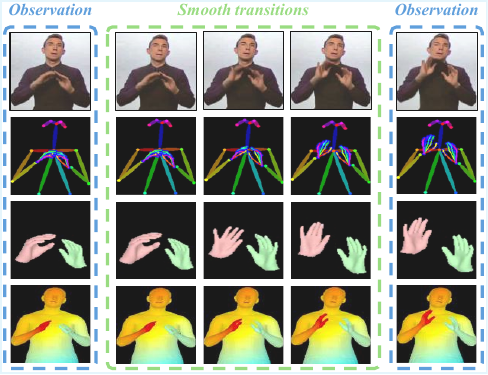}
  \caption{Visualization results generate a smooth transition from observation.}
  \label{fig: transition}
\end{figure}

\noindent\textbf{Exquisite finger details.} As shown in Table~\ref{ham}, we evaluate the generation results of hand details on the PHOENIX14T dataset from two aspects: semantics and video quality. Our method is highly competitive, achieving a BLEU-1 score of 30.53 and a ROUGE-F1 score of 30.49 on the Test set. The image quality indicators SSIM and PSNR are also better than SignAliger.

\noindent\textbf{Strong-3D full body.} As shown in Table~\ref{smp}, our method outperforms PTSLP and CogvideoX on both DEV and TEST sets, achieving 13.25 BLEU-4, slightly higher than SignAligner's 11.20, with SSIM 0.86 and PSNR 21.52, demonstrating stable and high-quality generation close to ground truth.

\noindent\textbf{Ultra semantics and clarity.} As shown in Table~\ref{video}, our method achieves state-of-the-art performance in sign language video generation, excelling in both semantic accuracy and visual quality. With the highest BLEU-1 (20.70/21.17) and BLEU-4 (8.45/8.55) scores on DEV/TEST sets, our approach better preserves linguistic content, while the strong SSIM (0.73) and PSNR (15.81/15.93) results demonstrate superior video clarity, outperforming existing methods in generating realistic and semantically precise sign language videos.

\begin{figure}[t]
  \centering
  \includegraphics[width=1\columnwidth]{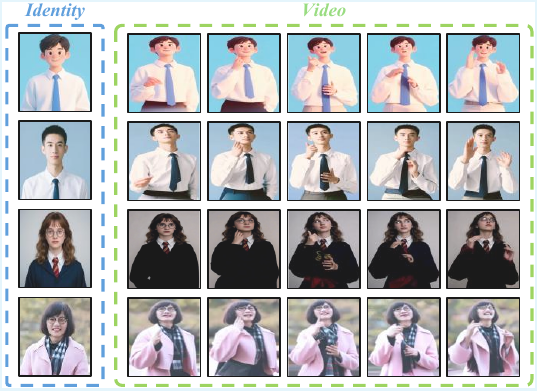}
  \caption{Visualization examples of sign language generation videos based on different identities.}
  \label{fig: identity}
\end{figure}

\subsection{Qualitative Results}
\noindent\textbf{Smooth motion transitions.} As shown in Figure~\ref{fig: transition}, we aim to generate smooth transitions between observations. We build on the initial observations (first and last columns), which primarily contain poses, hand details, and the 3D full body. Then, in the middle column, we successfully construct a smooth transition. Both the dynamics of the hand gesture and the adjustments to the hand joint positions show gradual changes. This demonstrates that our method can generate smooth transitions between observed actions, achieving a seamless transition from one observation to the next.

\noindent\textbf{Identity-aware video synthesis.} Figure~\ref{fig: identity} shows the synthesized sign language videos based on different identities. The identities represent representative figures of different genders and styles, and the videos generate corresponding sign language videos for these identities. We generate corresponding sign language content based on different identity information, intuitively showing how identity characteristics are presented in sign language, and providing an intuitive reference for the generation of personalized sign language videos.

\noindent\textbf{Visualization Results.} As shown in Figure~\ref{video}, we use text as input to first generate poses, hand details, and 3D full-body, and then synthesize the resulting sign language video. Compared to Ground Truth, the generated content closely matches the pose restoration, hand movement details, and full presentation, fully demonstrating the superior performance of our method in generating high-quality videos.

\section{Conclusion}
In this paper, we propose a novel paradigm for sign language video generation that decouples motion semantics from signer identity through a two-phase synthesis framework. We construct a multimodal sign lexicon that is independent of the signer. Subsequently, we leverage the motions in this lexicon to achieve motion synthesis from discrete to continuous motions. Finally, identity-aware video synthesis enables the generation of realistic videos of any signer. By treating motion as a portable "choreography layer," our approach overcomes the limitations of signer-specific datasets and enables unprecedented personalization flexibility. Experiments demonstrate that our approach not only achieves high semantic accuracy and visual quality but also bridges the "motion gap" between discrete sign motions, outperforming state-of-the-art methods on the PHOENIX14T dataset.

\bigskip
\bibliography{aaai2026}

\begin{thebibliography}{48}
\providecommand{\natexlab}[1]{#1}

\bibitem[{Ahn, Jang, and Chung(2024)}]{ahn2024slowfast}
Ahn, J.; Jang, Y.; and Chung, J.~S. 2024.
\newblock Slowfast Network for Continuous Sign Language Recognition.
\newblock In \emph{IEEE International Conference on Acoustics, Speech and Signal Processing}, 3920--3924.

\bibitem[{Cai et~al.(2023)Cai, Yin, Zeng, Wei, Sun, Yanjun, Pang, Mei, Zhang, Zhang et~al.}]{cai2023smpler}
Cai, Z.; Yin, W.; Zeng, A.; Wei, C.; Sun, Q.; Yanjun, W.; Pang, H.~E.; Mei, H.; Zhang, M.; Zhang, L.; et~al. 2023.
\newblock Smpler-X: Scaling up Expressive Human Pose and Shape Estimation.
\newblock \emph{Neural Information Processing Systems}, 11454--11468.

\bibitem[{Camgoz et~al.(2018{\natexlab{a}})Camgoz, Hadfield, Koller, Ney, and Bowden}]{Camgoz_Hadfield_Koller_Ney_Bowden_2018}
Camgoz, N.~C.; Hadfield, S.; Koller, O.; Ney, H.; and Bowden, R. 2018{\natexlab{a}}.
\newblock Neural Sign Language Translation.
\newblock In \emph{IEEE/CVF Conference on Computer Vision and Pattern Recognition}, 7784--7793.

\bibitem[{Camgoz et~al.(2018{\natexlab{b}})Camgoz, Hadfield, Koller, Ney, and Bowden}]{camgoz2018neural}
Camgoz, N.~C.; Hadfield, S.; Koller, O.; Ney, H.; and Bowden, R. 2018{\natexlab{b}}.
\newblock Neural Sign Language Translation.
\newblock In \emph{IEEE/CVF International Conference on Computer Vision}, 7784--7793.

\bibitem[{Chen et~al.(2022{\natexlab{a}})Chen, Wei, Sun, Wu, and Lin}]{Chen_2022_CVPR}
Chen, Y.; Wei, F.; Sun, X.; Wu, Z.; and Lin, S. 2022{\natexlab{a}}.
\newblock A Simple Multi-Modality Transfer Learning Baseline for Sign Language Translation.
\newblock In \emph{IEEE/CVF Conference on Computer Vision and Pattern Recognition (CVPR)}, 5120--5130.

\bibitem[{Chen et~al.(2022{\natexlab{b}})Chen, Zuo, Wei, Wu, Liu, and Mak}]{chen2022two}
Chen, Y.; Zuo, R.; Wei, F.; Wu, Y.; Liu, S.; and Mak, B. 2022{\natexlab{b}}.
\newblock Two-Stream Network for Sign Language Recognition and Translation.
\newblock \emph{NeurIPS}.

\bibitem[{Gong et~al.(2024)Gong, Foo, He, Rahmani, and Liu}]{gong2024llms}
Gong, J.; Foo, L.~G.; He, Y.; Rahmani, H.; and Liu, J. 2024.
\newblock Llms are Good Sign Language Translators.
\newblock In \emph{IEEE/CVF Conference on Computer Vision and Pattern Recognition}, 18362--18372.

\bibitem[{Gu et~al.(2022)Gu, Chen, Bao, Wen, Zhang, Chen, Yuan, and Guo}]{gu2022vector}
Gu, S.; Chen, D.; Bao, J.; Wen, F.; Zhang, B.; Chen, D.; Yuan, L.; and Guo, B. 2022.
\newblock Vector Quantized Diffusion Model for Text-to-Image Synthesis.
\newblock In \emph{IEEE/CVF Conference on Computer Vision and Pattern Recognition}, 10696--10706.

\bibitem[{Guo, Tang, and Wang(2019)}]{Guo_Tang_Wang}
Guo, D.; Tang, S.; and Wang, M. 2019.
\newblock Connectionist Temporal Modeling of Video and Language: A Joint Model for Translation and Sign Labeling.
\newblock In \emph{International Joint Conference on Artificial Intelligence}, 751--757.

\bibitem[{Ho, Jain, and Abbeel(2020)}]{ho2020denoising}
Ho, J.; Jain, A.; and Abbeel, P. 2020.
\newblock Denoising Diffusion Probabilistic Models.
\newblock \emph{Neural Information Processing Systems}, 6840--6851.

\bibitem[{Huang et~al.(2021)Huang, Pan, Zhao, and Tian}]{huang2021towards}
Huang, W.; Pan, W.; Zhao, Z.; and Tian, Q. 2021.
\newblock Towards Fast and High-Quality Sign Language Production.
\newblock In \emph{ACM International Conference on Multimedia}, 3172--3181.

\bibitem[{Jia et~al.(2025)Jia, Huang, Chen, Zhang, and Mao}]{jia2025d}
Jia, W.; Huang, M.; Chen, N.; Zhang, L.; and Mao, Z. 2025.
\newblock D\^{} 2iT: Dynamic Diffusion Transformer for Accurate Image Generation.
\newblock In \emph{IEEE/CVF Conference on Computer Vision and Pattern Recognition}, 12860--12870.

\bibitem[{Kingma and Ba(2015)}]{Kingma_Ba_2014}
Kingma, D.~P.; and Ba, J. 2015.
\newblock Adam: A Method for Stochastic Optimization.
\newblock In \emph{International Conference on Learning Representations}, 1--15.

\bibitem[{Mazumder et~al.(2021)Mazumder, Mukhopadhyay, Namboodiri, and Jawahar}]{Mazumder_Mukhopadhyay_Namboodiri_Jawahar_2021}
Mazumder, S.; Mukhopadhyay, R.; Namboodiri, V.~P.; and Jawahar, C. 2021.
\newblock Translating Sign Language Videos to Talking Faces.
\newblock In \emph{Indian Conference on Computer Vision, Graphics and Image Processing}, 1--10.

\bibitem[{McDonald et~al.(2016)McDonald, Wolfe, Schnepp, Hochgesang, Jamrozik, Stumbo, Berke, Bialek, and Thomas}]{McDonald_Wolfe_Schnepp_Hochgesang_Jamrozik_Stumbo_Berke_Bialek_Thomas_2016}
McDonald, J.; Wolfe, R.; Schnepp, J.; Hochgesang, J.; Jamrozik, D.~G.; Stumbo, M.; Berke, L.; Bialek, M.; and Thomas, F. 2016.
\newblock An Automated Technique for Real-Time Production of Lifelike Animations of American Sign Language.
\newblock \emph{Universal Access in the Information Society}, 551--566.

\bibitem[{Othman and Jemni(2011)}]{othman2011statistical}
Othman, A.; and Jemni, M. 2011.
\newblock Statistical Sign Language Machine Translation: from English Written Text to American Sign Language Gloss.
\newblock \emph{IJCSI}.

\bibitem[{Pavlakos et~al.(2024)Pavlakos, Shan, Radosavovic, Kanazawa, Fouhey, and Malik}]{pavlakos2024reconstructing}
Pavlakos, G.; Shan, D.; Radosavovic, I.; Kanazawa, A.; Fouhey, D.; and Malik, J. 2024.
\newblock Reconstructing Hands in 3D with Transformers.
\newblock In \emph{IEEE/CVF Conference on Computer Vision and Pattern Recognition}, 9826--9836.

\bibitem[{Peebles and Xie(2023)}]{peebles2023scalable}
Peebles, W.; and Xie, S. 2023.
\newblock Scalable Diffusion Models with Transformers.
\newblock In \emph{IEEE/CVF International Conference on Computer Vision}, 4195--4205.

\bibitem[{Rombach et~al.(2022)Rombach, Blattmann, Lorenz, Esser, and Ommer}]{rombach2022high}
Rombach, R.; Blattmann, A.; Lorenz, D.; Esser, P.; and Ommer, B. 2022.
\newblock High-Resolution Image Synthesis with Latent Diffusion Models.
\newblock In \emph{Computer Vision and Pattern Recognition}, 10684--10695.

\bibitem[{Sagawa and Takeuchi(2002)}]{sagawa2002teaching}
Sagawa, H.; and Takeuchi, M. 2002.
\newblock A Teaching System of Japanese Sign Language using Sign Language Recognition and Generation.
\newblock In \emph{ACM international conference on Multimedia}, 137--145.

\bibitem[{Saunders, Camgoz, and Bowden(2020)}]{saunders2020progressive}
Saunders, B.; Camgoz, N.~C.; and Bowden, R. 2020.
\newblock Progressive Transformers for End-to-End Sign Language Production.
\newblock In \emph{European Conference on Computer Vision}, 687--705. Springer.

\bibitem[{Saunders, Camgoz, and Bowden(2021)}]{saunders2021continuous}
Saunders, B.; Camgoz, N.~C.; and Bowden, R. 2021.
\newblock Continuous 3D Multi-Channel Sign Language Production via Progressive Transformers and Mixture Density Networks.
\newblock \emph{International journal of computer vision}, 2113--2135.

\bibitem[{Segouat(2009)}]{Segouat_2009}
Segouat, J. 2009.
\newblock A Study of Sign Language Coarticulation.
\newblock \emph{ACM Sigaccess Accessibility and Computing}, 31--38.

\bibitem[{Sohl-Dickstein et~al.(2015)Sohl-Dickstein, Weiss, Maheswaranathan, and Ganguli}]{Sohl-Dickstein_Weiss_Maheswaranathan_Ganguli_2015}
Sohl-Dickstein, J.; Weiss, E.; Maheswaranathan, N.; and Ganguli, S. 2015.
\newblock Deep Unsupervised Learning Using Nonequilibrium Thermodynamics.
\newblock In \emph{International Conference on Machine Learning}, 2256--2265.

\bibitem[{Song, Meng, and Ermon(2020)}]{Song_Meng_Ermon_2020}
Song, J.; Meng, C.; and Ermon, S. 2020.
\newblock Denoising Diffusion Implicit Models.
\newblock In \emph{International Conference on Learning Representations}.

\bibitem[{Stoll et~al.(2020)Stoll, Camgoz, Hadfield, and Bowden}]{stoll2020text2sign}
Stoll, S.; Camgoz, N.~C.; Hadfield, S.; and Bowden, R. 2020.
\newblock Text2Sign: Towards Sign Language Production using Neural Machine Translation and Generative Adversarial Networks.
\newblock \emph{International Journal of Computer Vision}, 891--908.

\bibitem[{Stoll, Hadfield, and Bowden(2020)}]{stoll2020signsynth}
Stoll, S.; Hadfield, S.; and Bowden, R. 2020.
\newblock Signsynth: Data-Driven Sign Language Video Generation.
\newblock In \emph{European Conference on Computer Vision}, 353--370.

\bibitem[{Tang et~al.(2022{\natexlab{a}})Tang, Guo, Hong, and Wang}]{tangslt2022}
Tang, S.; Guo, D.; Hong, R.; and Wang, M. 2022{\natexlab{a}}.
\newblock Graph-Based Multimodal Sequential Embedding for Sign Language Translation.
\newblock \emph{IEEE Transactions on Multimedia}, 4433--4445.

\bibitem[{Tang et~al.(2025{\natexlab{a}})Tang, He, Cheng, Wu, Guo, and Hong}]{tang2025discrete}
Tang, S.; He, J.; Cheng, L.; Wu, J.; Guo, D.; and Hong, R. 2025{\natexlab{a}}.
\newblock Discrete to Continuous: Generating Smooth Transition Poses from Sign Language Observations.
\newblock In \emph{Computer Vision and Pattern Recognition Conference}, 3481--3491.

\bibitem[{Tang et~al.(2025{\natexlab{b}})Tang, He, Guo, Wei, Li, and Hong}]{tang2025sign}
Tang, S.; He, J.; Guo, D.; Wei, Y.; Li, F.; and Hong, R. 2025{\natexlab{b}}.
\newblock Sign-IDD: Iconicity Disentangled Diffusion for Sign Language Production.
\newblock In \emph{AAAI Conference on Artificial Intelligence}, 7266--7274.

\bibitem[{Tang et~al.(2022{\natexlab{b}})Tang, Hong, Guo, and Wang}]{tang2022gloss}
Tang, S.; Hong, R.; Guo, D.; and Wang, M. 2022{\natexlab{b}}.
\newblock Gloss Semantic-Enhanced Network with Online Back-Translation for Sign Language Production.
\newblock In \emph{ACM International Conference on Multimedia}, 5630--5638.

\bibitem[{Tang et~al.(2025{\natexlab{c}})Tang, Xue, Wu, Wang, and Hong}]{tang2025gloss}
Tang, S.; Xue, F.; Wu, J.; Wang, S.; and Hong, R. 2025{\natexlab{c}}.
\newblock Gloss-Driven Conditional Diffusion Models for Sign Language Production.
\newblock \emph{ACM Transactions on Multimedia Computing, Communications and Applications}, 1--17.

\bibitem[{Vaswani et~al.(2017)Vaswani, Shazeer, Parmar, Uszkoreit, Jones, Gomez, Kaiser, and Polosukhin}]{vaswani2017attention}
Vaswani, A.; Shazeer, N.; Parmar, N.; Uszkoreit, J.; Jones, L.; Gomez, A.~N.; Kaiser, {\L}.; and Polosukhin, I. 2017.
\newblock Attention is All You Need.
\newblock \emph{Neural Information Processing Systems}, 30.

\bibitem[{Wang et~al.(2025{\natexlab{a}})Wang, Tang, Cheng, Li, Wang, and Hong}]{wang2025signaligner}
Wang, X.; Tang, S.; Cheng, L.; Li, F.; Wang, S.; and Hong, R. 2025{\natexlab{a}}.
\newblock SignAligner: Harmonizing Complementary Pose Modalities for Coherent Sign Language Generation.
\newblock \emph{arXiv preprint arXiv:2506.11621}.

\bibitem[{Wang et~al.(2025{\natexlab{b}})Wang, Tang, Song, Wang, Guo, and Hong}]{wangxu2025}
Wang, X.; Tang, S.; Song, P.; Wang, S.; Guo, D.; and Hong, R. 2025{\natexlab{b}}.
\newblock Linguistics-Vision Monotonic Consistent Network for Sign Language Production.
\newblock In \emph{IEEE International Conference on Acoustics, Speech and Signal Processing}, 1--5.

\bibitem[{Wang et~al.(2025{\natexlab{c}})Wang, Chen, Ma, Zhou, Huang, Wang, Yang, He, Yu, Yang et~al.}]{wang2025lavie}
Wang, Y.; Chen, X.; Ma, X.; Zhou, S.; Huang, Z.; Wang, Y.; Yang, C.; He, Y.; Yu, J.; Yang, P.; et~al. 2025{\natexlab{c}}.
\newblock Lavie: High-Quality Video Generation with Cascaded Latent Diffusion Models.
\newblock \emph{International Journal of Computer Vision}, 3059--3078.

\bibitem[{Wang, Gao et~al.(2002)}]{wang2002method}
Wang, Z.; Gao, W.; et~al. 2002.
\newblock A Method to Synthesize Chinese Sign Language Based on Virtual Human Technologies.
\newblock \emph{Journal of Software}, 2051--2056.

\bibitem[{Xie et~al.(2024)Xie, Zhang, Taiying, Tang, Du, and Li}]{xie2024g2p}
Xie, P.; Zhang, Q.; Taiying, P.; Tang, H.; Du, Y.; and Li, Z. 2024.
\newblock G2P-DDM: Generating Sign Pose Sequence from Gloss Sequence with Discrete Diffusion Model.
\newblock In \emph{AAAI Conference on Artificial Intelligence}, 6234--6242.

\bibitem[{Xue et~al.(2023)Xue, Liu, Yan, Zhou, Yuan, and Guo}]{xue2023alleviating}
Xue, W.; Liu, J.; Yan, S.; Zhou, Y.; Yuan, T.; and Guo, Q. 2023.
\newblock Alleviating Data Insufficiency for Chinese Sign Language Recognition.
\newblock \emph{Visual Intelligence}, 26.

\bibitem[{Yang et~al.(2024)Yang, Teng, Zheng, Ding, Huang, Xu, Yang, Hong, Zhang, Feng et~al.}]{yang2024cogvideox}
Yang, Z.; Teng, J.; Zheng, W.; Ding, M.; Huang, S.; Xu, J.; Yang, Y.; Hong, W.; Zhang, X.; Feng, G.; et~al. 2024.
\newblock Cogvideox: Text-to-Video Diffusion Models with an Expert Transformer.
\newblock \emph{arXiv preprint arXiv:2408.06072}.

\bibitem[{Yang et~al.(2023)Yang, Zeng, Yuan, and Li}]{yang2023effective}
Yang, Z.; Zeng, A.; Yuan, C.; and Li, Y. 2023.
\newblock Effective Whole-Body Pose Estimation with Two-Stages Distillation.
\newblock In \emph{IEEE/CVF International Conference on Computer Vision}, 4210--4220.

\bibitem[{Yin et~al.(2024)Yin, Li, Shen, Tang, and Zhuang}]{yin2024t2s}
Yin, A.; Li, H.; Shen, K.; Tang, S.; and Zhuang, Y. 2024.
\newblock T2S-GPT: Dynamic Vector Quantization for Autoregressive Sign Language Production from Text.
\newblock In \emph{Association for Computational Linguistics}, 3345--3356.

\bibitem[{Yue, Liao, and Loy(2025)}]{yue2025arbitrary}
Yue, Z.; Liao, K.; and Loy, C.~C. 2025.
\newblock Arbitrary-Steps Image Super-Resolution via Diffusion Inversion.
\newblock In \emph{IEEE/CVF International Conference on Computer Vision}, 23153--23163.

\bibitem[{Zhang et~al.(2025)Zhang, You, Shi, and Gu}]{zhang2025uncertainty}
Zhang, L.; You, W.; Shi, K.; and Gu, S. 2025.
\newblock Uncertainty-Guided Perturbation for Image Super-Resolution Diffusion Model.
\newblock In \emph{IEEE/CVF International Conference on Computer Vision}, 17980--17989.

\bibitem[{Zhao et~al.(2023)Zhao, Hu, Zhou, Shi, and Li}]{zhao2023best}
Zhao, W.; Hu, H.; Zhou, W.; Shi, J.; and Li, H. 2023.
\newblock BEST: BERT Pre-Training for Sign Language Recognition with Coupling Tokenization.
\newblock In \emph{AAAI Conference on Artificial Intelligence}, 3597--3605.

\bibitem[{Zhou et~al.(2023)Zhou, Chen, Clap{\'e}s, Wan, Liang, Escalera, Lei, and Zhang}]{zhou2023gloss}
Zhou, B.; Chen, Z.; Clap{\'e}s, A.; Wan, J.; Liang, Y.; Escalera, S.; Lei, Z.; and Zhang, D. 2023.
\newblock Gloss-Free Sign Language Translation: Improving from Visual-Language Pretraining.
\newblock In \emph{IEEE/CVF International Conference on Computer Vision}, 20871--20881.

\bibitem[{Zhou et~al.(2024)Zhou, Wang, Chen, Bai, Li, Zhang, Xu, Yang, and Wang}]{zhou2024realisdanceequipcontrollablecharacter}
Zhou, J.; Wang, B.; Chen, W.; Bai, J.; Li, D.; Zhang, A.; Xu, H.; Yang, M.; and Wang, F. 2024.
\newblock RealisDance: Equip Controllable Character Animation with Realistic Hands.

\bibitem[{Zuo et~al.(2024)Zuo, Wei, Chen, Mak, Yang, and Tong}]{zuo2024simple}
Zuo, R.; Wei, F.; Chen, Z.; Mak, B.; Yang, J.; and Tong, X. 2024.
\newblock A Simple Baseline for Spoken Language to Sign Language Translation with 3D Avatars.
\newblock In \emph{European Conference on Computer Vision}, 36--54.

\end{thebibliography}


\end{document}